\documentclass[conference]{IEEEtran}
\usepackage{graphicx}
\usepackage{booktabs}
\usepackage{amsmath,amssymb}
\usepackage{algorithm}
\usepackage{algpseudocode}
\usepackage{url}
\usepackage{cite}
\usepackage[hidelinks]{hyperref}

\title{CRISP: Corneal Confocal Microscopy Real-Time Image Stitching Pipeline}
\author{
\IEEEauthorblockN{Qincheng Qiao}
\IEEEauthorblockA{Department of Endocrinology and Metabolism\\
Qilu Hospital, Shandong University\\
Jinan 250012, China\\
Email: jugking6688@gmail.com}
\and
\IEEEauthorblockN{Puli Zhang}
\IEEEauthorblockA{Department of Endocrinology and Metabolism\\
Shanghai Sixth People's Hospital Affiliated to\\
Shanghai Jiao Tong University School of Medicine\\
Shanghai, China\\
Email: zhangpl21@163.com}
\and
\IEEEauthorblockN{Jian Zhou}
\IEEEauthorblockA{Department of Endocrinology and Metabolism\\
Shanghai Sixth People's Hospital Affiliated to\\
Shanghai Jiao Tong University School of Medicine\\
Shanghai, China\\
Email: zhoujian@sjtu.edu.cn}
\and
\IEEEauthorblockN{Xinguo Hou}
\IEEEauthorblockA{Department of Endocrinology and Metabolism\\
Qilu Hospital, Shandong University\\
Jinan 250012, China\\
Email: houxinguo@sdu.edu.cn}
}
\begin{document}
\maketitle
\begin{abstract}
Morphology of the sub-basal nerve plexus (SNP) reflects peripheral nerve health, and corneal confocal microscopy (CCM) provides an important means for in vivo, real-time, non-invasive observation of the SNP. However, mainstream CCM devices offer a limited field of view per frame, whereas the SNP is spatially non-uniform; discrete image sampling is therefore sensitive to sampling location and frame selection, which limits the reproducibility and clinical adoption of CCM as a quantitative assessment tool. Wide-field stitching can reconstruct larger SNP mosaics by integrating sequentially acquired CCM images, but existing methods largely rely on offline post-processing, additional hardware, or specific acquisition protocols, and lack open-source real-time solutions for conventional CCM video streams.

This paper presents CRISP (Corneal confocal microscopy Real-time Image Stitching Pipeline), an open-source real-time SNP wide-field stitching framework for conventional CCM examination video streams. CRISP excludes defocused and discontinuous segments via focus-aware gating, propagates poses through local pairwise registration, and maintains non-redundant spatial coverage with a sparse anchor map; when local temporal continuity is interrupted, the system completes relocalization and subgraph merging through global appearance retrieval followed by geometric verification. The framework prioritizes low-latency coverage feedback during examination while outputting accepted frames, poses, and anchor information to initialize offline fine stitching.

To our knowledge, CRISP is the first open-source real-time SNP wide-field stitching framework released for conventional CCM video streams. By lowering the barrier to adoption and reproduction of wide-field stitching, CRISP may help move SNP wide-field imaging from a research tool into routine clinical examination workflows. The CRISP project repository is available at \url{https://github.com/SummerColdWind/CRISP}.
\end{abstract}

\section{Introduction}

The adage that “the eye is the window to the soul” takes on a more concrete meaning in clinical medicine: the cornea allows direct observation of neural structures in living subjects \cite{al-aqaba2019Cornealnerveshealth}. Current clinical evidence indicates that SNP morphology can sensitively and reliably reflect the health status or severity of systemic peripheral neuropathy \cite{cruzat2017VivoConfocalMicroscopy}. CCM is presently the only commonly used clinical modality that enables real-time, non-invasive in vivo imaging of the SNP, and holds considerable promise for diagnosing and monitoring peripheral neuropathy \cite{chiang2023Invivocornealconfocal,qiao2025DevelopmentDiagnosticNomograms}.

Nevertheless, CCM has not yet become an approved surrogate endpoint biomarker, and limited scan area is a key factor constraining its clinical adoption \cite{petropoulos2018DiagnosingDiabeticNeuropathy}. The field of view of mainstream CCM devices is $400\times400~\mu\mathrm{m}$, covering only approximately 0.2\% of the total SNP area \cite{zhang2022DefiningOptimalSample}. Because the SNP is spatially non-uniform \cite{petropoulos2015InferiorWhorlDetecting,patel2005MappingNormalHuman}, fragmented image sampling introduces substantial diagnostic variability \cite{lagali2018Widefieldcornealsubbasal,badian2021Widefieldmosaicscorneal}, thereby limiting the reproducibility and scalability of CCM as a quantitative assessment modality.

Wide-field stitching registers sequentially acquired CCM images into a common coordinate frame to reconstruct SNP mosaics covering larger areas \cite{patel2005MappingNormalHuman,lagali2018Widefieldcornealsubbasal}, enabling sampling and quantitative analysis at relatively fixed anatomical locations, reducing subjective frame selection, and improving result consistency \cite{kheirkhah2015ComparisonStandardWideField,badian2021Widefieldmosaicscorneal,sandvik2026novelmethodstandardised}. Over the past two decades, CCM-based SNP wide-field stitching has evolved from manual assembly \cite{patel2005MappingNormalHuman,turuwhenua2012FullyAutomatedMontaging} to fully automated registration \cite{turuwhenua2012FullyAutomatedMontaging,vaishnav2017Rapidautomatedmosaicking}, and from offline processing \cite{turuwhenua2012FullyAutomatedMontaging,li2022NerveStitcherCornealconfocal} to real-time imaging \cite{zhivov2010Realtimemappingsubepithelial,allgeier2022Realtimelargeareaimaging,petroll2015VivoConfocalMicroscopy}. However, SNP wide-field stitching has still not been widely adopted in clinical practice \cite{sandvik2026novelmethodstandardised}.

\begin{figure*}[!t]
\centering
\includegraphics[width=\textwidth]{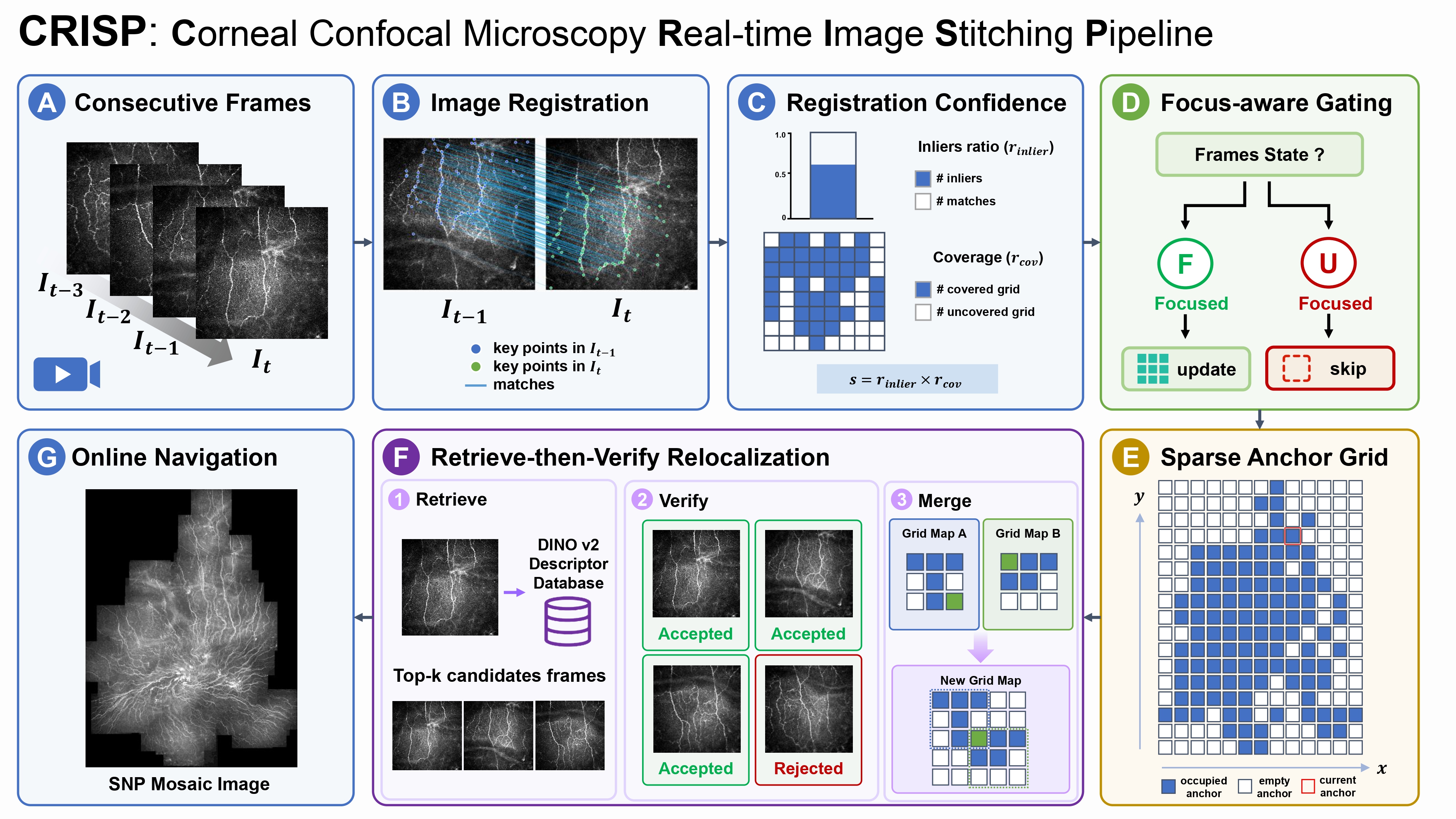}
\caption{Overview of the CRISP framework. (A) Continuous CCM frame sequence; (B) local feature matching and pairwise frame registration; (C) registration confidence $s=r_{\mathrm{inlier}}\cdot r_{\mathrm{cov}}$ based on RANSAC inlier ratio $r_{\mathrm{inlier}}$ and spatial coverage $r_{\mathrm{cov}}$; (D) focus-aware gating that classifies acquisition state as focused (F) or unfocused (U) according to geometric continuity between adjacent frames; (E) sparse anchor grid, with blue occupied anchors and the red box indicating the anchor corresponding to the current frame; (F) retrieve-then-verify relocalization: Top-$K$ candidates are recalled from a DINOv2 descriptor database, accepted or rejected after local geometric verification, and grids are merged when reliable geometry is established across subgraphs; (G) online navigation output: SNP wide-field mosaic.}
\label{fig:crisp-overview}
\end{figure*}

This paper presents CRISP (Corneal confocal microscopy Real-time Image Stitching Pipeline), a real-time wide-field stitching framework for conventional CCM video streams (Figure~\ref{fig:crisp-overview}). CRISP combines local registration, global retrieval, and sparse spatial constraints to incrementally maintain a navigation map that guides acquisition in real time. CRISP requires no additional hardware or specialized acquisition protocol: given a conventional CCM examination video stream, it performs fully automatic stitching and provides real-time feedback and guidance to the operator. CRISP can also serve as a front end for offline fine stitching by supplying a sufficient set of non-redundant accepted frames and initial poses.

\section{Methods}

\subsection{Problem Definition and System Overview}

Given a grayscale video stream obtained in real time during a conventional corneal confocal microscopy (CCM) examination, let the input image sequence be $\{I_t\}_{t=1}^{T}$, where $I_t\in\mathbb{R}^{H\times W}$ denotes the $t$-th frame. Because a single CCM frame covers only a local region of the sub-basal nerve plexus (SNP), the goal of wide-field stitching is to incrementally register valid frames into a unified two-dimensional coordinate system and provide real-time feedback on covered regions during acquisition. CRISP (Corneal confocal microscopy Real-time Image Stitching Pipeline) formulates this task as online sparse mapping: the system selects from the raw video stream a set of valid frames $\mathcal{F}$ that are geometrically reliable and spatially non-redundant, estimates their map-center coordinates $\{\mathbf{p}_i\}_{I_i\in\mathcal{F}}$ with $\mathbf{p}_i=(x_i,y_i)^\top\in\mathbb{R}^2$, and maintains a sparse coverage map $\mathcal{M}$ for real-time navigation.

CRISP targets low-latency feedback during examination rather than offline global optimization after acquisition. The system must therefore satisfy three constraints: invalid frames caused by defocus, blinking, or hand tremor should be prevented from entering the map; highly overlapping consecutive frames should be sparsified to avoid redundant accumulation; and when a clear acquisition segment resumes after interruption, the current frame should be re-associated with the existing map. Accordingly, CRISP comprises three interchangeable modules: (1) a local registration module for estimating pairwise geometry and reliability scores; (2) a global retrieval module for recalling potentially overlapping reference frames from historical valid frames; and (3) a sparse anchor map module that writes verified frames to regular spatial anchors to form a low-redundancy coverage representation.

For motion modeling, CRISP approximates dominant inter-frame motion with two-dimensional translation. For query frame $I_q$ and reference frame $I_r$, if local registration yields relative translation $\Delta\mathbf{p}_{r\rightarrow q}$, the query pose is updated as

$$
\mathbf{p}_q=\mathbf{p}_r-\Delta\mathbf{p}_{r\rightarrow q}.
$$

This translational assumption is consistent with the primary probe motion along the corneal tangent plane during conventional CCM examination and enables online map updates at modest computational cost.

\subsection{Local Registration Module}

The local registration module supplies unified geometric evidence for adjacent-frame continuity assessment, pose propagation, retrieval-candidate verification, and subgraph merging. Given frame pair $(I_a,I_b)$, the system first obtains a candidate match set through a local feature extractor and matcher:

$$
\mathcal{C}_{ab}=\{(\mathbf{u}_j,\mathbf{v}_j)\}_{j=1}^{N},
$$

where $\mathbf{u}_j$ and $\mathbf{v}_j$ are the coordinates of matched points in the two frames. Because CCM images exhibit repeated nerve structures, sparse local texture, and uneven contrast, relying solely on match count is vulnerable to locally clustered false correspondences. CRISP therefore defines local registration confidence as the product of the RANSAC geometric inlier ratio and the spatial coverage of inliers:

$$
s(I_a,I_b)=r_{\mathrm{inlier}}\cdot r_{\mathrm{cov}}.
$$

Here,

$$
r_{\mathrm{inlier}}=\frac{|\mathcal{C}_{ab}^{\mathrm{in}}|}{|\mathcal{C}_{ab}|}
$$

denotes the RANSAC inlier ratio; if the number of candidate matches is insufficient or geometric estimation fails, $s(I_a,I_b)=0$. Spatial coverage $r_{\mathrm{cov}}$ is computed by partitioning the estimated overlap region into a regular grid and counting the fraction of grid cells that contain inliers:

$$
r_{\mathrm{cov}}=\frac{|\Omega_{\mathrm{occ}}|}{|\Omega|}.
$$

This score requires matches to be both geometrically consistent and spatially dispersed across the overlap region.

After a frame pair passes geometric verification, the system estimates two-dimensional translation using the median displacement of matched points:

$$
\Delta\mathbf{p}_{a\rightarrow b}
=
\operatorname{median}\{\mathbf{v}_j-\mathbf{u}_j\}_{j=1}^{N},
$$

with medians computed separately along the $x$ and $y$ axes. This estimator is robust to a small number of outlier matches and avoids costly online global optimization.

\subsection{Global Retrieval Module}

Local registration is well suited to short continuous segments, but after blinking, defocus, or revisiting a prior path, the immediately preceding frame is often an unreliable reference. The global retrieval module rapidly recalls a small set of candidates from historical valid frames to support relocalization and connection across acquisition segments. For each map-accepted valid frame $I_i\in\mathcal{F}$, the system extracts an image-level descriptor with global appearance encoder $\phi(\cdot)$:

$$
\mathbf{f}_i=\phi(I_i)\in\mathbb{R}^{d}.
$$

For query frame $I_q$, the system computes descriptor $\mathbf{f}_q$ and retrieves historical frames by cosine similarity:

$$
\operatorname{sim}(I_q,I_i)
=
\frac{\mathbf{f}_q^\top\mathbf{f}_i}{\|\mathbf{f}_q\|_2\|\mathbf{f}_i\|_2}.
$$

Global retrieval is responsible only for candidate recall. CRISP adopts a retrieve-then-verify strategy: local geometric verification is applied only to the Top-$K$ candidates whose similarity exceeds a threshold; a candidate frame is accepted as a reference only when $s(I_i,I_q)$ exceeds the geometric threshold. This design avoids pairwise registration against the entire history while preserving strict geometric consistency.

The module addresses two principal scenarios. First, when the system re-enters the focused state from a non-focused state, the current frame is attached to the existing map via retrieval; if retrieval fails, a new local subgraph is started. Second, during continuous focused acquisition, the system may retrieve candidates both within the current subgraph and across subgraphs: the former supplies more stable historical references, whereas the latter discovers spatial overlap between subgraphs and triggers merging.

\subsection{Sparse Anchor Map Module}

The sparse anchor map module converts a continuous video stream into a low-redundancy spatial coverage representation. Because adjacent CCM frames are typically highly overlapping, writing every valid frame into the map would substantially increase storage, retrieval, and visualization cost. CRISP maintains a regular anchor grid in map coordinates, with at most one representative frame per anchor. For anchor $a_k$ in subgraph $\mathcal{G}^{(m)}$, the discrete coordinates are $(r_k,c_k)$ and the continuous position is

$$
\mathbf{q}_k=(g c_k,g r_k)^\top,
$$

where $g$ is the anchor spacing. For frame $I_i$ with an estimated pose, the system queries its $K_a$ nearest anchors and computes

$$
d_{ik}=\|\mathbf{p}_i-\mathbf{q}_k\|_2.
$$

If the current frame is closer to the anchor center than the observation already bound to a candidate anchor and satisfies the minimum spatial separation constraint, it is bound to that anchor; otherwise it is rejected. This mechanism retains only spatially informative representative observations and suppresses redundant sampling in highly overlapping regions.

Whenever binding succeeds, the system generates the eight neighboring anchors of that anchor so that the map expands incrementally along the acquisition path. CRISP also permits multiple local subgraphs. When refocus after defocus cannot relocalize to an existing subgraph, the system either retains the current subgraph and starts a new one, or discards short segments with insufficient coverage. If frames in two subgraphs establish reliable geometry through cross-subgraph retrieval and local verification, the system computes the inter-subgraph translational offset, transforms all occupied anchor frames in the source subgraph, and rebinds them to the target subgraph to complete merging.

\subsection{Online System Workflow}

\subsubsection{Focus-aware frame gating}

CRISP maintains a binary state $z_t\in\{\mathrm{Focused},\mathrm{Unfocused}\}$, denoting focused and unfocused acquisition, respectively. The state is driven by local registration confidence between adjacent frames: if $s(I_{t-1},I_t)$ exceeds a threshold, the system considers the current segment geometrically continuous and in focus; otherwise the temporal chain is deemed broken. Transition from unfocused to focused requires several consecutive frames satisfying this condition, or successful local verification between the current frame and the most recent focused valid frame. Once adjacent-frame verification fails in the focused state, the system immediately returns to unfocused and suspends map updates.

\subsubsection{Online map update and relocalization}

In the focused state, the system selects a historical reference within the current subgraph or a candidate reference from global retrieval, performs local verification on the current frame, estimates relative translation, and updates its map pose. If the current frame is successfully bound to an anchor, it is added to the valid frame set and indexed for global retrieval; if binding fails, its spatial information is already sufficiently represented by nearby anchors and it is not stored as a new map observation. When recovery from unfocused state cannot find a reliable reference in the existing map, the system creates a new subgraph or removes invalid short subgraphs. If the current frame is simultaneously verified against a candidate in another subgraph, subgraph merging is executed.

\subsubsection{Real-time visualization and output}

After processing each frame, CRISP generates a sparse navigation map from the frames and poses associated with occupied anchors in the current subgraph. This output provides coverage feedback during acquisition rather than replacing offline fine stitching. After the examination, the system retains valid frames, anchor coordinates, subgraph relations, and initial poses, which can initialize subsequent high-precision mosaic reconstruction and quantitative analysis.

\begin{algorithm}[htbp]
\caption{Online CRISP stitching}
\label{alg:online-crisp}
\begin{algorithmic}[1]
\Require Streaming CCM frames $\{I_t\}$
\Ensure Sparse map $\mathcal{M}$, accepted frames $\mathcal{F}$, poses $\{\mathbf{p}_i\}$
\State Initialize sparse anchor map $\mathcal{M}$, frame database $\mathcal{F}$, retrieval index $R$
\State Set acquisition state $z \gets \mathrm{Unfocused}$
\State Set previous frame $I_{\mathrm{prev}} \gets \mathrm{None}$
\For{each incoming frame $I_t$}
    \State Extract local features for $I_t$
    \If{$I_{\mathrm{prev}}$ is None}
        \State $I_{\mathrm{prev}} \gets I_t$
        \State \textbf{continue}
    \EndIf
    \State Compute local confidence $s(I_{\mathrm{prev}}, I_t)$
    \State Update $z$ using temporal geometric consistency
    \If{$z = \mathrm{Focused}$}
        \State Retrieve candidate references from $R$ when needed
        \State Verify candidates by local registration
        \If{a valid reference $I_r$ is found}
            \State Estimate translation $\Delta\mathbf{p}_{r\rightarrow t}$
            \State Update pose $\mathbf{p}_t \gets \mathbf{p}_r - \Delta\mathbf{p}_{r\rightarrow t}$
            \State Bind $I_t$ to the sparse anchor map
            \If{binding is accepted}
                \State Add $I_t$ to $\mathcal{F}$ and update $R$
            \EndIf
            \State Merge grids if a cross-grid reference is verified
        \Else
            \State Start a new grid or discard an invalid short grid
        \EndIf
    \EndIf
    \State Render the current sparse map for real-time feedback
    \State $I_{\mathrm{prev}} \gets I_t$
\EndFor
\end{algorithmic}
\end{algorithm}

\subsection{Implementation Details}

In our implementation, input frames are $384\times384$ grayscale images. Local registration uses SuperPoint \cite{detone2018SuperPointSelfSupervisedInteresta} for keypoint extraction and LightGlue \cite{lindenberger2023LightGlueLocalFeaturea} for matching, retaining at most 512 keypoints per frame. Geometric verification employs RANSAC to estimate a homography and obtain the inlier set; spatial coverage is computed on an $8\times8$ grid; pose propagation uses the median displacement of matched points as a two-dimensional translation. The local geometric threshold is $\tau_m=0.5$.

Global retrieval uses DINOv2 ViT-S/14 \cite{oquab2024DINOv2LearningRobust} as the image-level encoder. Input images are converted to three channels, resized to $224\times224$, and normalized with ImageNet statistics, yielding 384-dimensional descriptors. The retrieval similarity threshold is $\tau_s=0.8$. Top-5 historical candidates are retrieved for relocalization; during focused acquisition, the numbers of within-subgraph and cross-subgraph candidates are 3 and 2, respectively.

Anchor spacing is $g=W/3$ with $W=384$. Each binding query considers the $K_a=4$ nearest anchors, and the minimum spatial separation is $W/9$. A new subgraph is initialized with a central anchor and its eight neighbors; after successful binding, eight neighbors are generated to expand the map. If the maximum side length of the bounding box of covered area in a subgraph reaches 450 pixels, that subgraph is retained after interruption; otherwise it is treated as an invalid short segment and removed. Main hyperparameters are summarized in Table~\ref{tbl:hyperparams}.

\begin{table}[htbp]
\centering
\caption{Main hyperparameters}
\label{tbl:hyperparams}
\begin{tabular}{llr}
\toprule
Parameter & Meaning & Value \\
\midrule
$W$ & Input frame side length & 384 \\
$\tau_m$ & Local geometric threshold & 0.5 \\
$\tau_s$ & Retrieval similarity threshold & 0.8 \\
$N_c$ & Consecutive focused frames & 3 \\
$g$ & Anchor spacing & $W/3$ \\
$K_a$ & Number of anchor neighbors & 4 \\
$d_{\min}$ & Minimum binding separation & $W/9$ \\
$K$ & Relocalization candidates & 5 \\
$K_{\mathrm{in}}$ & Within-subgraph candidates & 3 \\
$K_{\mathrm{out}}$ & Cross-subgraph candidates & 2 \\
$L_{\min}$ & Minimum valid subgraph extent & 450 px \\
\bottomrule
\end{tabular}
\end{table}

\section{Experiments and Results}

\begin{figure*}[!t]
\centering
\includegraphics[width=\textwidth]{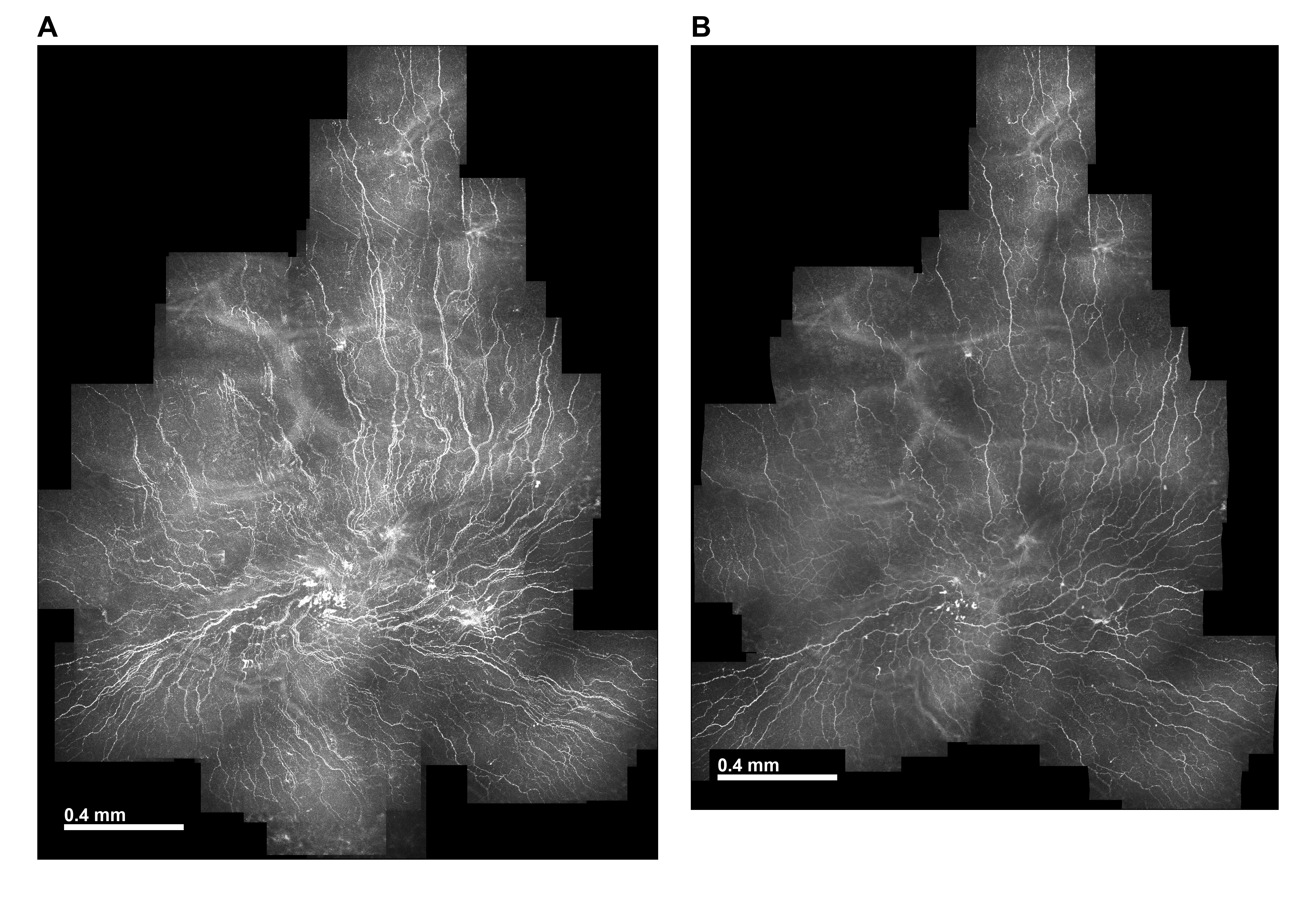}
\caption{Wide-field SNP mosaic of the right eye in one subject. (A) CRISP online stitching result covering the inferior whorl and central vertically oriented nerve region; image size $1997\times2619$~px, area approximately 3.63~mm$^2$ (about 22.68$\times$ single-frame field of view); (B) mosaic refined offline with HIT using CRISP accepted frames and initial poses as input; size $1891\times2459$~px, area approximately 3.25~mm$^2$ (about 20.3$\times$ single-frame field of view). Scale bar: 0.4~mm.}
\label{fig:mosaic-result}
\end{figure*}

\subsection{Experimental Setup}

We performed CCM examination on one subject, Xinguo Hou, a co-author of this paper. Right-eye CCM imaging was conducted with the Rostock Cornea Module connected to a Heidelberg Retina Tomograph III (Heidelberg Engineering, Heidelberg, Germany). The CRISP system ran on an external laptop (Intel(R) Core(TM) Ultra 9 275HX CPU and NVIDIA GeForce RTX 5090 Laptop GPU) connected via Ethernet to the PC attached to the CCM device.

Before examination, topical anesthesia was administered with proparacaine hydrochloride. A drop of carbomer gel was placed at the center of the objective as coupling medium, and a sterile corneal cap was mounted on the objective. The participant was instructed to keep the head stable, face the microscope objective, and maintain forward gaze throughout the examination. The operator gradually advanced the objective until the corneal cap gently contacted the central cornea. After the corneal epithelium was visualized, focus was adjusted to image the sub-basal nerve plexus. During the examination, the operator monitored CRISP output on the laptop in real time to assess the current mosaic and guide further scanning.

\subsection{Experimental Results}

The total examination lasted approximately 10 minutes and successfully captured the subject's inferior whorl region of spirally distributed SNP together with the centrally located vertically oriented SNP region above it (Figure~\ref{fig:mosaic-result}). The mosaic produced automatically by CRISP measured $1997\times2619$~px, corresponding to an actual SNP area of 3.6295~mm$^2$, or 22.68 times the area of a single field of view.

The single-frame set corresponding to this mosaic was subsequently processed with HIT (HRT Imaging Tool) \cite{allgeier2026HITfullyautomated}. The resulting image measured $1891\times2459$~px, corresponding to an actual SNP area of 3.2485~mm$^2$, or 20.30 times the single-field area. HIT further reduced nonlinear registration errors present after coarse alignment, yielding a mosaic more suitable for subsequent quantitative analysis or other evaluation. The slightly smaller output size may reflect differences in frame-selection protocols.

\section{Discussion}

Existing SNP wide-field stitching approaches face major barriers to practical deployment and reproducible adoption. On one hand, some real-time or large-area imaging solutions depend heavily on additional hardware, customized device modifications, or specific acquisition protocols, making them difficult to deploy directly in conventional CCM examination settings \cite{allgeier2014MosaickingSubbasalNerve,allgeier20183Dconfocallaserscanning,allgeier2022Realtimelargeareaimaging}. On the other hand, apart from a few studies \cite{li2022NerveStitcherCornealconfocal,qiao2023NerveStitcher20EvolutionStitching,allgeier2026HITfullyautomated}, most methods do not provide complete open-source implementations, which limits algorithm reproduction and integration into real examination workflows by clinical researchers. For clinically oriented SNP wide-field stitching, robustness and generality under conventional devices, conventional video streams, and conventional operating conditions are central determinants of practical value.

Several open-source offline stitching tools already exist, including two versions of NerveStitcher and HIT \cite{li2022NerveStitcherCornealconfocal,qiao2023NerveStitcher20EvolutionStitching,allgeier2026HITfullyautomated}. These tools provide usable methods for post-examination mosaic reconstruction, but offline pipelines cannot report current coverage during acquisition \cite{allgeier2022Realtimelargeareaimaging}. Operators must ensure sufficient scan coverage and continuous inter-frame overlap during examination; if the acquisition trajectory contains gaps, repetitions, or blind spots, the final mosaic may appear fragmented. Relying on offline stitching alone therefore does not resolve coverage guidance during acquisition.

Real-time SNP wide-field stitching faces two prominent challenges. First, CCM is contact-based imaging with shallow depth of field \cite{petroll2015VivoConfocalMicroscopy,allgeier2014MosaickingSubbasalNerve}; involuntary eye movement, blinking, and operator hand tremor can cause defocus and disrupt continuous imaging \cite{allgeier2014MosaickingSubbasalNerve,allgeier2022Realtimelargeareaimaging}. Second, complete SNP mosaics typically require hundreds to thousands of valid images \cite{patel2005MappingNormalHuman,turuwhenua2012FullyAutomatedMontaging}, and registration time against historical frames grows with the number of acquired frames, limiting real-time usability of wide-field stitching \cite{allgeier2026HITfullyautomated}.

To address these issues, we present CRISP, an open-source real-time SNP wide-field stitching framework for conventional CCM video streams. Results from one real CCM examination show that CRISP produced an online navigation map spanning approximately 22.68 single-frame fields of view, indicating that the framework can reconstruct large SNP coverage without additional hardware. Moreover, accepted frames, poses, and anchor information from CRISP can initialize offline fine stitching; processing CRISP coarse registration with HIT \cite{allgeier2026HITfullyautomated} yields a more refined SNP mosaic. CRISP therefore supports both real-time acquisition navigation and serves as a front end for offline stitching software, linking online coverage feedback with post-acquisition fine reconstruction.

CRISP is not tied to any single image registration algorithm. Local registration and global retrieval are relatively independent modules; the current release implements them with SuperPoint \cite{detone2018SuperPointSelfSupervisedInteresta}, LightGlue \cite{lindenberger2023LightGlueLocalFeaturea}, and DINOv2 \cite{oquab2024DINOv2LearningRobust}. These components are not fixed assumptions of the framework; as the field advances, CRISP can replace registration or retrieval backends while preserving the overall pipeline. This plug-in design facilitates rapid adoption of improved algorithms in SNP wide-field stitching applications.

This study has several limitations. First, experiments were conducted on only a single case, which is insufficient to characterize performance across devices, subject conditions, and operator habits. Larger-scale evaluation is needed to assess coverage, relocalization stability, and runtime efficiency. Second, the local registration and global retrieval methods used here were not designed specifically for SNP images; general-purpose vision models may produce mismatches or retrieval bias on this modality \cite{li2022NerveStitcherCornealconfocal,chen2026Robustregistrationlarge}. Training or fine-tuning feature extraction and matching networks tailored to SNP images may further improve CRISP robustness. Finally, this work is currently released as a preliminary study on a preprint platform.

\bibliographystyle{plain}
\bibliography{refs}
\end{document}